\pdfoutput=1
\documentclass[letterpaper]{article}
\usepackage{iccc}
\usepackage{times}
\usepackage{amsmath}
\usepackage{graphicx}
\usepackage{booktabs}

\title{Recipes for Creativity: Iterative Generation and Evaluation in Large Language Models}
\author{
Rens Anderson \\
LIACS \\
Leiden University \\
Einsteinweg 55 \\
2333 CC Leiden \\
The Netherlands
\And
Tessa Verhoef \\
LIACS \\
Leiden University \\
Einsteinweg 55 \\
2333 CC Leiden \\
The Netherlands
\And
Amirhossein (Miros) Zohrehvand \\ 
LIACS \\
Leiden University \\
Einsteinweg 55 \\
2333 CC Leiden \\
The Netherlands\\
}

\begin{document}
\maketitle

\begin{abstract}
Generative models are often evaluated through singular artifacts, whereas human creativity typically emerges through iterative generation, appraisal, and refinement. This pilot study examines whether iterative search improves LLM creativity by adapting FunSearch to recipe generation for the 2024 Pillsbury Bake-Off and evaluating outputs against human benchmarks using TTCT-based LLM evaluation. Across two experiments, we test iteration count, generator temperature, and in-loop selection-scorer model size. Results show that iterative generation-selection can produce recipes with creativity scores comparable to human benchmarks, but additional iterations alone do not improve creativity. The in-loop evaluator matters most: a smaller selection scorer yields significantly higher scores across most TTCT dimensions, while temperature has limited effects except for originality. These findings suggest that evaluator design is a first-order design variable in subjective creative search.
\end{abstract}

\section{Introduction}

In recent years, Generative Artificial Intelligence (GenAI), and Large Language Models (LLMs) in particular, have gained significant attention for their ability to generate creative content across various domains \cite{genai_creativity,llm_creativity_content,artificial_muses,ai_more_creative_divergent,stevenson_aut_2022}. Research in computational creativity with such models often focuses on individual creative artifacts as isolated system outputs, whereas human creative activity is rarely characterized by such discreteness. In practice, creative output frequently emerges through an iterative process of generation, evaluation, and refinement \cite{iterative_creativity,characterising_creative_process}. 

In this study, we investigate LLM creative capabilities through iterative generation strategies. We leverage the FunSearch algorithm \cite{funsearch}, which has shown strong performance in objective tasks such as solving mathematical problems. FunSearch is an evolutionary search procedure in which an LLM is repeatedly prompted with high-scoring prior candidates from a programs database, partitioned into semi-isolated islands for diversity, and proposes new candidates that an automatic scorer admits or discards, biasing generation toward better-scoring outputs without fine-tuning. FunSearch is especially relevant to computational creativity because it searches over generative procedures rather than isolated outputs, making it suitable for domains where reusable, inspectable procedures matter. We extend this line of inquiry by applying it to a creative problem where output quality is inherently subjective: developing new cooking recipes for the Pillsbury Bake-Off. We particularly focus on model size \cite{larger_llm_v2} and sampling temperature \cite{temperature_creativity_parameter}. \textbf{RQ1} asks whether iterative FunSearch improves on a near one-shot baseline and approaches a human benchmark. \textbf{RQ2} asks which of three design factors most strongly shapes creativity scores: iteration count, generator temperature, or in-loop selection-scorer size. We conduct two experiments to address these questions. 

This paper makes two main contributions to the literature on generative AI for creative artifact production. First, it extends FunSearch, originally developed for mathematical and other codified objective tasks, to a subjective creative domain by applying it to recipe generation for the Pillsbury Bake-Off. This provides a concrete computational-creativity test case for studying iterative generation-selection loops beyond standard optimization benchmarks. Second, it provides an empirical evaluation of which parts of that iterative pipeline matter most for creative performance. By comparing iteration count, generator temperature, and in-loop selection-scorer model size, we show that selection pressure exerted by the evaluator is more consequential than simply increasing the number of search cycles or changing sampling temperature.

\section{Background}

Computational creativity has considered creativity not as a property of an isolated artifact, but as a relation between artifact, generator, and evaluative context. For example, frameworks such as DIFI locate creativity in the interaction among domain, individual, and field \cite{csikszentmihalyi1988society,DIFI,boden_creativity_def}. A FunSearch-style loop naturally distributes these roles: the generator proposes candidates, the programs database preserves lineages, and the in-loop scorer acts as a gatekeeping field.

\begin{figure*}[ht]
  \centering
  \includegraphics[width=0.85\textwidth]{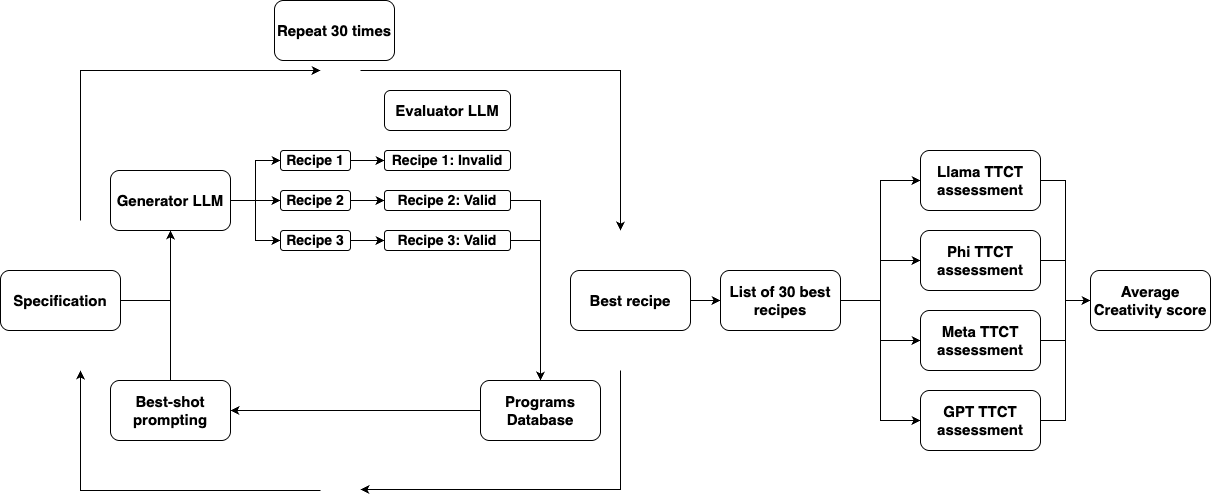}
  \caption{Conceptual framework illustrating how the FunSearch algorithm is integrated with TTCT-derived evaluation to assess the creativity of generated recipe artifacts.}
  \label{fig:creativity_framework}
\end{figure*}

FunSearch was introduced for mathematical and other codified tasks where progress can be measured against objective criteria \cite{funsearch}. Its design links it to evolutionary computation, where semi-isolated subpopulations help preserve diversity and prevent premature convergence \cite{harada_alba_parallel_ga,gong_distributed_ea_survey}. This is appealing for computational creativity because variation and selective retention are central to both evolutionary views of creativity and iterative accounts of human creative work \cite{characterising_creative_process,mexica_engagement_reflection}.

This approach differs from other iterative LLM creativity methods. Systems such as Self-Refine critique and revise a single draft \cite{self_refine}, while multi-critic and beam-style approaches improve outputs through comparative feedback and broader search over candidate continuations \cite{collective_critics,creative_beam_search,pushing_gpt_creativity_limits}. FunSearch is closer to the latter family: in our adaptation, it improves outputs by retaining and reusing strong candidates rather than editing one evolving recipe.

Evaluation is the hardest part of this transfer. Human judgment remains the gold standard for creative artifacts \cite{human_golden}, and computational-creativity research warns that evaluation methods must match the creativity claim being made \cite{lamb_evaluating_cc_2018}. At the same time, recent work suggests that structured LLM-based evaluation can be useful for creativity-related tasks when carefully prompted, even though concerns remain about evaluator overlap and self-preference bias \cite{Assessing_creativity_LLM_TTCT,llm_agree_creativity,self_preference_bias,LLM_Evaluators_Prefer_Themselves}.

\section{Study Design and Evaluation}

We adapt FunSearch to the Pillsbury Bake-Off by defining a recipe ``skeleton'' from the competition rules and using a multi-island search process. Each island maintains a local candidate set, and later prompts are built from high-scoring recipes in that island using best-shot prompting \cite{funsearch}. The system therefore improves through guided re-generation from selected examples, rather than by directly editing a single draft.

The recipe skeleton operationalized the Bake-Off constraints directly. Candidate artifacts required a title, a bounded ingredient list, exactly one official Pillsbury ingredient, concise preparation instructions, a short preparation time, and an accompanying story. Each run also maintained multiple semi-isolated islands with local candidate pools rather than one global database. This supported diversity: islands could preserve different recipe lineages and explore different parts of the search space, even with the same model family throughout the pipeline. A full workflow diagram of this adapted pipeline is provided in Figure~\ref{fig:creativity_framework}, showing how the generation loop, in-loop selection scorer, and post hoc TTCT-style evaluators fit together. The in-loop scorer applies the Pillsbury-style rubric during search to decide which candidates survive into future prompts. The four TTCT-style evaluators are applied only after generation, to the retained best recipe from each run, and are used only for the analyses reported below.

During generation, an in-loop \textit{selection scorer} applies a Pillsbury-style rubric to rank candidate recipes. This search objective uses the competition's weighted logic of recipe plus accompanying story, because both were generated during search. Within each run, top candidates are retained according to this score across islands; when multiple recipes tie for the highest score, one is selected at random. Final creativity is assessed separately after generation by four LLM evaluators using TTCT-derived dimensions: fluency, flexibility, originality, and elaboration \cite{torrance_ttct_1966,Assessing_creativity_LLM_TTCT}. We follow prior work that adapts TTCT-style dimensions to products rather than treating them as direct psychometric measures of an agent's creative capacity \cite{art_or_artifice}.

Here, the TTCT dimensions are interpreted at the level of the recipe as a product. Fluency refers to the perceived richness of ideas within a single recipe, flexibility to the variety of conceptual or culinary perspectives it combines, originality to the novelty or uncommonness of the concept, and elaboration to the amount of detail and development in the final artifact. We therefore measure how generated recipes are judged under a structured creativity lens, not the model's latent creative capacity.

The evaluation prompts were designed to stabilize judgment rather than stimulate generation. Evaluators provided a brief qualitative rationale before assigning a score, using a chain-of-thought style structure to separate explanation from final rating \cite{chain_of_thought_google,chain_of_thought}. Persona framing was held fixed as a prompt-design choice rather than varied experimentally. This keeps the claims tied to the three manipulated factors rather than to every prompt detail.

For partial calibration, we compare generated recipes with a human reference set: the 2024 Pillsbury Bake-Off winning recipe plus 29 randomly selected entries from the same competition. Because stories for the human entries were not public, the final benchmark uses only the recipe component. This gives us a useful reference set, but not a complete ground-truth measure of recipe creativity.

This creates an asymmetry between generation-time and comparison-time evaluation. During search, recipes and stories were jointly optimized because the Bake-Off rubric rewards both; during final comparison, only recipe content could be retained. The benchmark should therefore be read as partial calibration, not a fully matched contest replication. In practice, the benchmark check placed the human winning recipe near the top under both candidate scorers rather than at rank one, indicating some sensitivity to quality but also clear divergence from the official human outcome.

We report two experiments. Experiment 1 varies iteration count, comparing a one-iteration near one-shot baseline with 5, 15, and 30 iterations. Experiment 2 varies generator temperature and in-loop scorer size in a balanced $3 \times 2$ design. Temperature matters because prior work links it to novelty--coherence trade-offs in creative language generation \cite{temperature_creativity_parameter}. Scorer size matters because larger models often perform better on objective tasks, but that relationship may not transfer cleanly to subjective creative search. In the full implementation, the generator was based on Meta-17B, while in-loop scoring used either Meta-8B or Meta-17B. Experiment 2 fixed seven islands and a batch size of five so that differences could be attributed more directly to temperature and scorer size than to search breadth.

Before the main comparisons, we checked whether the model-based scorer could rank the human reference set in a plausible way. The benchmark did not reproduce the official competition outcome, but it provided a useful calibration point: the winning human recipe remained a strong performer, while the mismatch supports our caution that LLM evaluation is only a proxy for human judgment.

\section{Results and Interpretation}

\begin{figure}[t]
  \centering
  \includegraphics[width=\columnwidth]{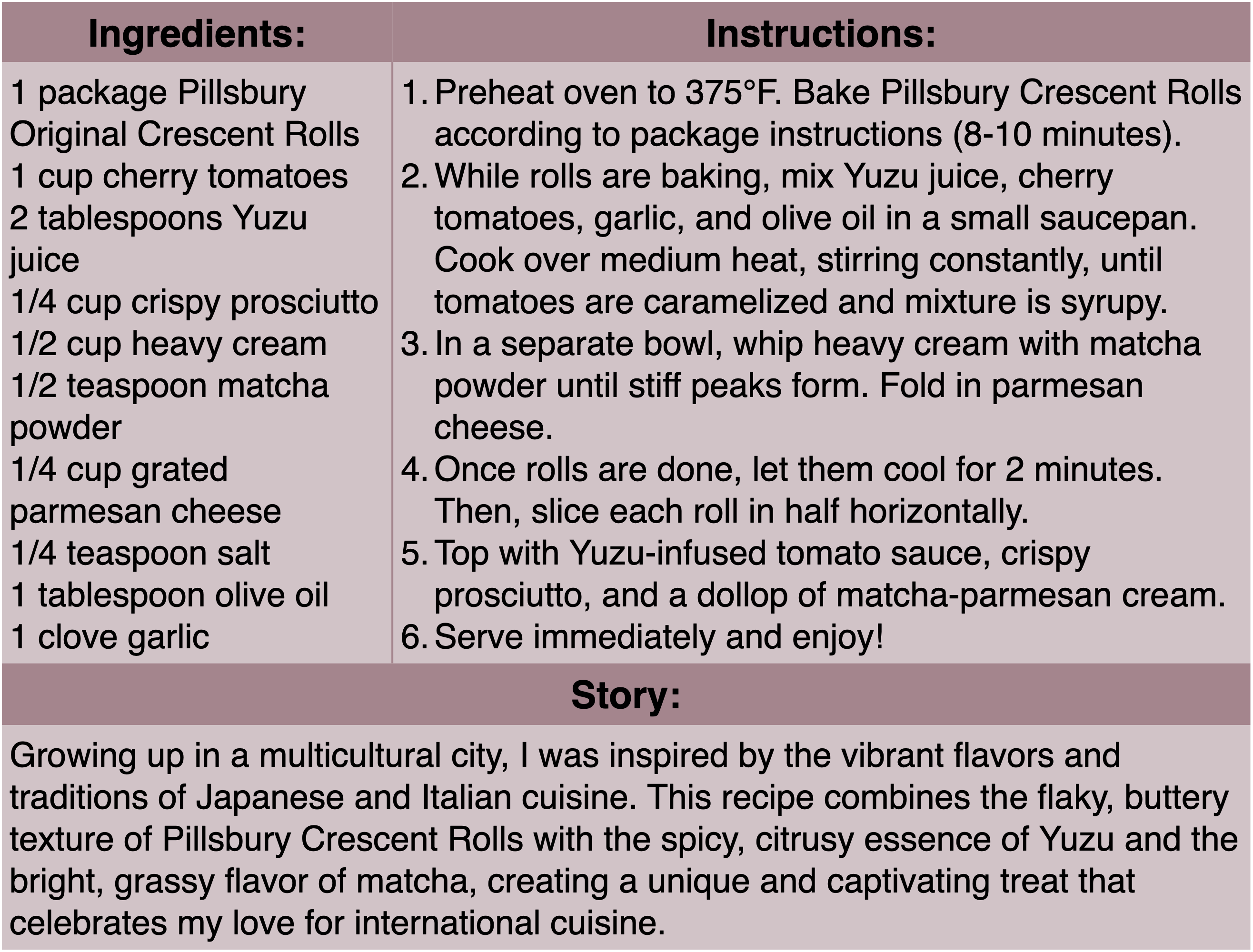}
  \caption{Example recipe generated by the LLM after $30$ iterations.}
\label{fig:recipe}
  \vspace{1mm}
  {\noindent\footnotesize\raggedright\textit{Notes:} Compared with the Pillsbury reference set, artifacts of this kind tend to combine more unusual flavor pairings (e.g.\ truffle with breakfast staples, prosciutto-wrapped asparagus) while remaining within the skeleton's structural constraints.\par}
\end{figure}

Experiment 1 asks whether more search helps. Figure~\ref{fig:recipe} shows a representative recipe retained after 30 iterations of FunSearch with the Meta-17B generator (the appendix provides the recipe skeleton, the in-loop rubric, the evaluation prompt structure). We find more search helps only weakly. Iteration counts of 5, 15, and 30 produced mean final creativity scores of 3.921, 3.835, and 3.927, respectively, compared with 3.638 for the Pillsbury reference set. This suggests that the iterative setup can reach human-comparable scores under the evaluation used here, but does not show a clear monotonic gain from adding cycles.

\begin{figure*}[ht]
  \centering
  \includegraphics[width=\textwidth]{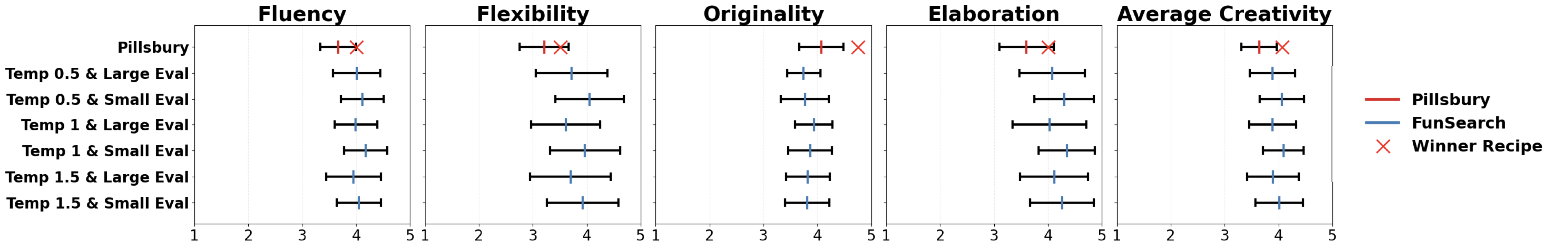}
  \caption{Average creativity scores across TTCT dimensions for FunSearch-generated recipes under each experimental condition, compared to the Pillsbury Bake-Off baseline.}
  \label{fig:creativity_results}
  \vspace{1mm}
  {\noindent\footnotesize\raggedright\textit{Notes:} Scores range from $1$ (low) to $5$ (high). Each row represents a configuration defined by generator temperature (Temp: $0.5$, $1.0$, or $1.5$) and selection-scorer model (Large = $Meta-17B$, Small = $Meta-8B$). The red cross marks the winning Pillsbury recipe. Error bars indicate one standard deviation.\par}
\end{figure*}

The one-iteration baseline sharpens this interpretation. Removing repeated search substantially lowered the in-loop weighted recipe score, from 4.71 to 4.00, while leaving the final creativity score nearly unchanged at roughly 4.1. This is unsurprising considering the in-loop evaluator uses other criteria and does not optimize for creativity.

Experiment 1 also found the FunSearch conditions had larger standard deviations than the human reference set on average (e.g.\ SD $= 0.411$ at 30 iterations versus $0.328$ for Pillsbury). One explanation could be that search explores a broader and less uniform solution space than the benchmark recipes through this iterative process. 

Experiment 2 varies generator temperature and in-loop evaluator size. As presented in Figure ~\ref{fig:creativity_results}, the smaller 8B selection scorer yields higher scores than the 17B scorer across average creativity, fluency, flexibility, and elaboration. One possible explanation is that in subjective search, larger evaluators may not be better aligned with the kinds of novelty and variation rewarded by the final assessment framework. Further, Figure \ref{fig:creativity_results} shows temperature effects are weaker. Lower generator temperature reduces originality, consistent with the intuition that less randomness leads to more predictable outputs \cite{temperature_creativity_parameter}. Higher temperatures, however, do not significantly improve any TTCT dimension. Sampling diversity alone therefore does not guarantee stronger creative artifacts, especially when later stages privilege coherence, structure, and in-loop scoreability.

We also conduct regression analyses to explore this effect further (for full results please see the appendix Table~\ref{tab:regression}). Relative to the $1.5$ temperature and Meta-8B reference condition, the larger in-loop scorer showed significant negative associations with overall creativity, fluency, flexibility, and elaboration, while originality was unaffected. Temperature effects were mostly negligible apart from the drop in originality at the lowest temperature. Adjusted $R^2$ values were low across models, meaning that temperature and scorer size capture only part of what drives final judgments. Other factors may include the specific examples retained within islands, prompt framing effects, and evaluator heterogeneity. 

Taken together, the two experiments indicate that evaluator design exerts stronger influence on final scores than either added search depth or added stochasticity alone. From a computational-creativity perspective, this shifts the question from whether an LLM can generate many novel candidates to what kind of design makes iterative novelty accumulate in subjective settings. If the in-loop evaluator rewards artifacts that are well formed but conservative, repeated search may stabilize around competent but unsurprising recipes. A looser evaluator might broaden exploration at the cost of coherence and practical plausibility. Future systems should therefore vary not only temperature or iteration count, but also the diversity, architecture, and incentive structure of evaluators inside the loop. That design space is especially relevant for subjective creative domains where no single oracle exists.

\section{Limitations and Future Work}

Our setup uses the FunSearch algorithm, a program-search framework that has previously been applied only to objective tasks, to generate recipes for the Pillsbury Bake-Off. Within the algorithm, the loop behaves less like direct refinement and more like selective re-generation. This may explain why iteration count alone has little effect on final creativity scores. If each step samples a new candidate from high-scoring examples instead of revising one draft, search may stay near already acceptable recipes rather than showing a clear trajectory of creative improvement. 

We believe that recipe generation is a useful testbed, though, for computational-creativity research on iterative search because it is neither fully open nor fully objective. A successful recipe must be novel, but also coherent, plausible, and attractive as an artifact meant to be cooked and judged. This makes it a bridge case between divergent tasks, where quantity and novelty can dominate, and objective tasks, where correctness can dominate.

The main limitation is that the study relies on LLM-based evaluation at both the in-loop and final-analysis stages. Using four evaluators reduces dependence on any one model's idiosyncrasies, but does not solve alignment with human judgment. The benchmark check showed only partial agreement with the human competition outcome, and that agreement was not uniform across dimensions: the human winning recipe, for example, received a higher originality assessment than the generated recipes. Accordingly, this paper can say more about how outputs behave under LLM evaluation than about how people would judge them as creative culinary artifacts.

Further, Meta-17B also served as both the in-loop scorer and as one of the final evaluators, introducing methodological overlap. Averaging across four models softens but does not remove this dependence between optimization and outcome assessment. In addition, TTCT-style LLM scoring is also a partial proxy for culinary creativity. Future work should combine human evaluation, evaluator diversity, and trajectory-level analysis of how recipes change across iterations.

\section{Conclusion}
This short paper presents an initial computational-creativity study of FunSearch in a subjective artifact domain. The results suggest that iterative search can produce recipes competitive with a human reference set under LLM-based evaluation, but that evaluator design matters more than additional search depth, and that lower temperature mainly reduces originality. The broader contribution is a case study in what changes when iterative search moves from objective optimization to subjective creative evaluation. The core methodological claim is that, in subjective creative search, the in-loop evaluator should be chosen very carefully and play an important role in the optimization of iterated generation set-ups. Future creative LLM systems may benefit less from ``more generation'' alone than from more deliberate design of the evaluative ecology: the diversity, architecture, and incentive structure of the scorers that decide which candidates survive. 

\clearpage
\begingroup
\small
\bibliographystyle{iccc}
\bibliography{iccc}
\endgroup

\clearpage
\onecolumn
\appendix
\section*{Appendix}

This appendix contains the conceptual pipeline, recipe skeleton, in-loop Pillsbury-style rubric, shared evaluation prompt structure, and full regression output from Experiment 2. 

\subsection{A.\ Recipe Skeleton}

The recipe skeleton operationalizes the 2024 Pillsbury Bake-Off rules as structural constraints on every candidate artifact. A candidate is retained in its island's programs database only if it satisfies all of the following:

\begin{itemize}
  \item A recipe title.
  \item At most ten ingredients, excluding staple items such as salt and pepper.
  \item Exactly one ingredient from the official Pillsbury ingredient list.
  \item A stated preparation time under 30 minutes.
  \item Preparation instructions of at most 2000 characters.
  \item An accompanying story of at most 500 characters.
\end{itemize}

Candidates that violate any constraint are discarded before scoring. The story is part of the artifact during generation because it enters the in-loop rubric; it is excluded from the final benchmark comparison because the public human reference set did not include matching stories.

\subsection{B.\ In-Loop Selection Rubric}

During search, each valid candidate is scored by the in-loop selection scorer using a weighted Pillsbury-style rubric: recipe content contributes 70\% of the score and the accompanying story contributes 30\%. Each sub-criterion is rated on a 1--5 Likert scale, and the weighted sum is rescaled to the 1--5 range.

\begin{itemize}
  \item \textbf{Recipe component (70\%):} taste, appearance, creativity, crowd appeal.
  \item \textbf{Story component (30\%):} narrative connection to the recipe, expression of family values or traditions, expression of personal passion.
\end{itemize}

This rubric is used only inside the search loop, not for the final creativity analysis. It is distinct from the TTCT-derived evaluation in the results: the rubric proxies competition judging, whereas the TTCT dimensions proxy product-level creativity.

\subsection*{C.\ Evaluation Prompt Structure}

Both the in-loop rubric prompts and the post hoc TTCT prompts share three fixed design elements, held constant across all conditions so that results can be attributed to the manipulated factors rather than prompt choices. Each prompt (i) assigns a persona (Pillsbury Bake-Off participant or competition judge, depending on the task), (ii) asks the model to produce a short qualitative rationale \emph{before} assigning a numerical score, and (iii) constrains the final output to a fixed JSON schema so that scores can be parsed reliably without regex-brittle postprocessing. The schematic structure of every evaluation prompt is:

\begin{quote}
\textit{[Persona]} You are a Pillsbury Bake-Off \{participant $\mid$ judge\}.\\
\textit{[Task]} Evaluate the following recipe artifact along \{dimension\_name\}, defined as \{product-level definition\}.\\
\textit{[Reasoning]} First write a short rationale grounded in the artifact's content.\\
\textit{[Scoring]} Then return a score on a 1--5 Likert scale.\\
\textit{[Format]} Return a single JSON object with fields \texttt{rationale} and \texttt{score}.
\end{quote}

For the TTCT prompts, the product-level definitions are: \emph{fluency} = perceived richness of ideas within the single recipe and its framing; \emph{flexibility} = variety of conceptual or culinary perspectives combined in the recipe; \emph{originality} = novelty or uncommonness of the recipe concept; \emph{elaboration} = amount of concrete detail and development in the final artifact. Persona framing was held fixed rather than varied experimentally, so the paper makes no causal claim about its effect.

\subsection*{D.\ Experiment 2: Full Regression Output}

Table~\ref{tab:regression} reports coefficient estimates from five dummy-coded OLS models predicting the overall creativity score and each TTCT dimension separately. Temperature $1.5$ with the Meta-8B selection scorer is the reference condition, so coefficients are differences from that baseline. The single consistent effect is the negative coefficient on \textit{LargeEvaluator} across four of five outcomes; temperature shows a significant effect only on originality, and only at the lowest temperature.

\begin{table}[!t]
\caption{Coefficient estimates from dummy-coded OLS models predicting creativity dimensions.}
\label{tab:regression}
\centering
\small
\setlength{\tabcolsep}{4pt}
    \begin{tabular}{lccccc}
    \toprule
     & (1) & (2) & (3) & (4) & (5) \\
    Variable & Creativity & Fluency & Flexibility & Originality & Elaboration \\
    \midrule
    LowTemp     & -0.017 & -0.021 &  0.096 & -0.144* &  0.002 \\
                & (0.079) & (0.079) & (0.124) & (0.072) & (0.111) \\
    MidTemp     & -0.037 & -0.083 &  0.023 & -0.083  & -0.002 \\
                & (0.079) & (0.079) & (0.124) & (0.072) & (0.111) \\
    LargeEvaluator   & -0.161* & -0.128* & -0.307* &  0.021  & -0.231* \\
                & (0.065) & (0.065) & (0.101) & (0.059) & (0.090) \\
    Constant     & 4.070  & 4.147  & 3.941  & 3.890   & 4.303 \\
                & (0.065) & (0.065) & (0.101) & (0.059) & (0.090) \\
    \textit{N}   & 180    & 180    & 180    & 180     & 180 \\
    Adj $R^2$    & 0.019  & 0.012  & 0.037  & 0.006   & 0.019 \\
    \bottomrule
    \end{tabular}\\[2mm]
\parbox{0.98\textwidth}{\small\raggedright\textit{Notes:} The model is $\text{Creativity} = \alpha + \beta \cdot \text{MidTemp} + \gamma \cdot \text{LowTemp} + \delta \cdot \text{LargeEvaluator} + \varepsilon$, where temperature $1.5$ and the $Meta$-$8B$ selection scorer form the reference condition. \textit{MidTemp} equals $1$ when the generator temperature is $1.0$, and $0$ otherwise. \textit{LowTemp} equals $1$ when the generator temperature is $0.5$, and $0$ otherwise. \textit{LargeEvaluator} equals $1$ if the in-loop selection scorer is $Meta$-$17B$, and $0$ if it is $Meta$-$8B$. The outcome variable is one of the four divergent thinking scores (Fluency, Flexibility, Originality, and Elaboration) or their average (Creativity). Standard errors are in parentheses. $*\, p < 0.05$.}
\end{table}

\end{document}